\documentclass[letterpaper, 10pt, conference]{ieeeconf}      % Use this line for a4 paper

\IEEEoverridecommandlockouts                              % This command is only needed if 
\usepackage{graphics} % for pdf, bitmapped graphics files
\usepackage{epsfig} % for postscript graphics files
\usepackage{mathptmx} % assumes new font selection scheme installed
\usepackage{times} % assumes new font selection scheme installed
\usepackage{amsmath} % assumes amsmath package installed
\usepackage{amssymb}  % assumes amsmath package installed
\usepackage{lmodern}
\usepackage{color}
\usepackage{bm}
\newcommand{\argmax}{\mathop{\rm arg~max}\limits}
\newcommand{\argmin}{\mathop{\rm arg~min}\limits}

\title{\LARGE \bf
An Optimal Assistive Control Strategy \\ based on User's Motor Goal Estimation	
}

\author{Jun-ichiro Furukawa$^{1}$ and Jun Morimoto$^{1}$% <-this % stops a space
	\thanks{$^{1}$Dept. of Brain Robot Interface, ATR computational Neuroscience Labs}%
}

\begin{document}

\maketitle
\thispagestyle{empty}
\pagestyle{empty}
%%%%%%%%%%%%%%%%%%%%%%%%%%%%%%%%%%%%%%%%%%%%%%%%%%%%%%%%%%%%%%%%%%%%%%%%%%%%%%%%
\begin{abstract}
In this study, we propose an optimal assistive control strategy that uses estimated user's movement intention as the terminal cost function. We estimate the movement intention by observing human user's joint angle, angluar velocity, and muscle activities for very short period of time. A task-related low-dimensional feature space is extracted from the observed user's movement data. We assume that discrete number of optimal control laws associated to different target tasks are pre-computed. Then, the optimal assistive policy is derived by blending the pre-computed optimal control laws based on the linear Bellman combination method. Coefficients that determine how to blend the control laws are derived based on the low-dimensional feature value that represents the user's movement intention. To validate our proposed method, we conducted basketball throwing tasks. In these experiments, subjects were asked to throw a basketball into a hoop placed at different throwing distances. The distances from the throwing point to the hoop were estimated as the user's movement intention and the optimal control policies were derived by using our proposed method. The results showed that the basketball throwing performances of the subjects were mostly improved. 
\end{abstract}
%%%%%%%%%%%%%%%%%%%%%%%%%%%%%%%%%%%%%%%%%%%%%%%%%%%%%%%%%%%%%%%%%%%%%%%%%%%%%%%%
\section{INTRODUCTION}
Because of the recent progress in robotics technologies, wearable robots, such as exoskeleton robots, are expected to physically interact with and assist humans in their activities.
As proof-of-principle, the hand exoskeleton \cite{JimsonNgeo:2013tp, hand_exo2016}, and upper and lower-body exoskeleton robots \cite{furukawaICRA2015, Young2017, Gui2017} have been studied.
For these applications, the use of surface electromyography (EMG) can be a possible approach to intuitively control the robots by estimating the user's movement intention \cite{Kentaro2011, Panagiotis2010}.

Concretely, using linearly scaled electromyography (EMG) to derive joint torques of an assistive device 
is a standard approach for an exoskeleton robot control.
However, to accomplish a given task based on this simple EMG-based control strategy,
human users need to generate highly-tuned EMG profiles by themselves.
Therefore, this standard approach can increase the computational burden of human users nervous  system
even though it may be able to reduce the physical burden. 
The other popular approach to estimate human movement intention is applying a classification method
to estimate an associated movement to observed EMG signals.
In this case, EMG is only used for initiating pre-designed movements. Therefore, from computational point of view, 
it seems suitable approach for assistive control.
However, we can use this approach only for limited number of discrete target tasks
since all the assitive controllers associated to the target tasks need to be pre-computed.

%%%%%%%%%%%%%%%%%%%%%%%%%%%%%%%%%%%%%
% Figure 1
%%%%%%%%%%%%%%%%%%%%%%%%%%%%%%%%%%%%%
\begin{figure}[t]
	\includegraphics[width=80mm]{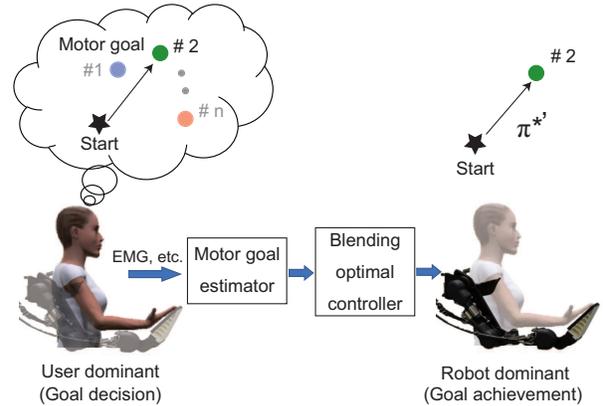}
	\begin{center}
		\caption{Schematic illustration of our approach. Optimal assistive policy is derived by blending pre-computed optimal control laws
based on the linear Bellman combination method. How to blend control laws are determined from estimated user's movement intention.}
		\label{fig:intro}
	\end{center}
\end{figure}
%%%%%%%%%%%%%%%%%%%%%%%%%%%%%%%%%%%%%

In this study, we develop an assistive control method that takes advantage of the above two standard approaches.
In our proposed method,        
We estimate the movement intention by observing human user's joint angle, angluar velocity, and muscle activities
for very short period of time. A low-dimensional feature space that is related a given task
is extracted from these observed data related to user's movements.  
We assume that discrete number of optimal control laws associated to different tasks are pre-computed.
Then, the optimal assistive policy is derived by blending the pre-computed optimal control laws
based on the linear Bellman combination method.
Coefficients that determine how to blend the control laws are derived by monitoring 
the low-dimensional feature value that represents the user's movement intention (see Fig. \ref{fig:intro}). 
To validate our proposed method, we conducted basketball throwing tasks. In these experiments, subjects were asked to throw a basketball into a hoop placed at different throwing distances. The distances from the throwing point to the hoop were estimated as the user's movement intention and the optimal control policies were derived by using our proposed method. The results showed that the basketball throwing performances of the subjects were mostly improved.

The rest of this paper is organized as follows.
In Section II, we have introduced the related EMG-based control strategies for the exoskeleton robot.
In Section III, we explain our user's goal estimation-based optimal control approach.
Section IV describes the experimental setups.
Section V describes our experimental results.
Finally, the conclusion is provided in Section VII.

\section{Related works}
%We briefly introduce the conventional EMG-based control strategy for the exoskeleton robot in this section.
In order to intuitively control assisitive devices such as exoskeleton robots, regression approaches are often used to estimate the user's movement intention.
In these approaches, a motion intention was estimated by finding relationships between EMG signals and joint torques  with linear or non-linear models \cite{Ao2017, furukawaTRO, Fleischer}.
In this method, the robot can be continuously controlled according to the monitored muscle activities.
However, since the fixed relationships were used in these regression methods, human users need to carefully 
generate own EMG profiles to cope with variety of tasks.

Another frequently used approach to estimate user's movement intention is using a classification method.
For example, support vector machine (SVM) \cite{Oskoei2008} or linear discriminant analysis (LDA) \cite{Zhang2014}
were adopted to classify measured information detected from human users.
The classification results were used to initiate pre-designed control output patterns associated to 
the estimated class labels.

In this study, we take the advantages of these two approaches.
We first prepare pre-optimaized assistive control laws.
Then these optimal control laws are combined using linear Bellman combination method\cite{LBC_silva}
to generate continuous optimal control output for exoskeleton robot.
Therefore, the human users do not need to carefully generate their own EMG signals since
our method derives optimal policy for a given task estimated from the measured data caused by user's movements.
Furthermore,  since we just combine the pre-optimized control laws to derive the corresponding optimal policy 
to the given task, the proposed approach is computationally light and can easily derive the policy in realtime.

\section{METHODS}
\begin{figure}[t]
	\includegraphics[width=80mm]{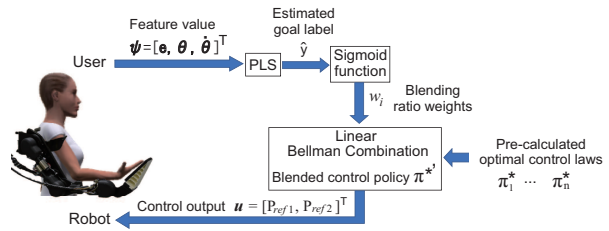}
	\begin{center}
		\caption{Proposed method}
		\label{fig:proposed_model}
	\end{center}
\end{figure}

This section introduces our approach for determining the control policy to assist the user's motor goal (see Fig. \ref{fig:proposed_model}).

%%%%%%%%%%%%%%%%%%%%%%%%%%%%%%%%%%%%%
% Figure 2
%%%%%%%%%%%%%%%%%%%%%%%%%%%%%%%%%%%%%
%%%%%%%%%%%%%%%%%%%%%%%%%%%%%%%%%%%%%

%%%%%%%%%%%%%%%%%
%% PLS
%%%%%%%%%%%%%%%%%
\subsection{User's motor goal estimation}
To cope with the continuous change of user's motor goal, we first find low-dimensional feature space
to determine how to combine the pre-optimized control outputs.

%In most conventional motor goal estimations, the estimator training is performed on the data sets of the sampled features and corresponding class labels with a classification approach, e.g., support vector machine (SVM) \cite{Oskoei2008}, and linear discriminant analysis (LDA) \cite{Zhang2014}.
%However, such methods are generally aimed at determining the class labels existing in the training data, and it is difficult to continuously estimate the changing classes that are not present in the training data. 

In this study, we used the partial least squares (PLS) algorithm \cite{PLS1, PLS2} to find the task related low-dimensional feature space.
First, the user's states ${\boldsymbol \psi} = \left[\xi_{1}, \xi_{2}, ..., \xi_{m}\right]^{\top}$
associated with the goal labels $y$ were observed.
Here we considered a dataset with $n$ samples as :
 ${\bf \Psi} = \left[{\boldsymbol \psi}_{1}, {\boldsymbol \psi}_{2},...,{\boldsymbol \psi}_{n}\right]^{\top}$, and ${\bf y} = \left[y_{1},y_{2},...,y_{n}\right]^{\top}$.
In PLS, for a given sample set, the direction ${\bf W} \in \mathcal{R}^{m \times j}$ of the $j$-dimensional feature space is determined such that the sample covariance of ${\boldsymbol \psi}_{i}$ and $y_{i}$ is maximized in the space of the feature as described below:
\begin{equation}
{\bf W} = \argmax_{||{\bf r}||=1}\left[cov({\bf \Psi}{\bf r},{\bf y})\right]^{2}.
\end{equation}
The projection from the original $m$-dimensional feature space to a $j$-dimensional space ($j < m$) can be performed as follows:
\begin{equation}
{\boldsymbol \mu}={\bf W}^{\top}{\boldsymbol \psi}
\end{equation}
As described above, since the subspace is selected in consideration with the covariance with the class labels, the low-dimensional features reflecting the label information can be obtained.
In this study, the user's motor goal was estimated directly by the regression model between the low-dimensional feature ${\boldsymbol \mu}$ and label $y$. The estimated $\hat{y}$ is used to determine the coefficients $w_{i}$ for the linear Bellman combination in (\ref{eq:combination}).
To map the $\hat{y}$ as blending ratio weights $w_{i}$ between $0$ and $1$, we use the sigmoid function as follows:
\begin{equation}\label{eq:weight}
w_{i} = \frac{1}{1+\exp(-a\hat{y}-b)}
\end{equation}
where, the $a (> 0)$ and $b$ are the parameters adjusted according to experimental conditions.

%%%%%%%%%%%%%%%%%%%%%%%%%%%%%
% linear bellman combination
%%%%%%%%%%%%%%%%%%%%%%%%%%%%%
\subsection{Optimal control based on motion intention}
In this study, we considered deriving the control policy according to the user's motor goal as follow: 
\begin{equation*}
\pi^{*'} = M(x,\psi),
\end{equation*}
where, $\pi^{*'}$ indicates the optimal control policy for achieving the target goal, $x$ indicates the state variable, and $\psi$ reflects the user's motor goal.
The purpose here is to find the optimal policy that achieves the user's motor goal.
A simple method is to set an objective function corresponding to the user’s target goal and optimize it.
However, creating a controller that achieves the target usually needs either time-consuming manual tuning dependent on experience or cost-intensive computational burden for the optimization.
In addition, it is difficult to create a controller corresponding to each target if the targets change continuously.
Therefore, presently, we considered reusing the optimal controllers which were obtained for the new target by combining them as follows:
\begin{equation}\label{eq:LBC}
\pi^{*'} = \alpha_{1}(x,t)\pi^{*}_{1}(x,t)+...+\alpha_{n}(x,t)\pi^{*}_{n}(x,t),
\end{equation}
where, $\alpha(x,t)$ is the mixing coefficient, and $\pi^{*}_{n}$ indicates the sub-optimal control policy for $n$ the component of the motor goal task obtained by minimizing the total cost (or objective function) $v(\cdot)$ as follows:
\begin{equation}\label{eq:cost_func}
v^{\pi}(x,t) = g(x(T))+\sum_t^{T-1} l(x,\pi,t)
\end{equation}
\begin{equation}
\pi^{*} \leftarrow \argmin_{\pi}v^{\pi}
\end{equation}
where, $g(\cdot)$ and $l(\cdot)$ indicate the terminal cost and instantaneous cost, respectively.

Assuming the availability of a collection of $n$ finite horizon control problems that share all setting except the terminal cost $g(x(T))$ of (\ref{eq:cost_func}), the linear bellman combination approach \cite{LBC_silva}, \cite{LBC_todorov} can be used, and the $\alpha(x,t)$ can be written as follows:
\begin{equation}\label{eq:combination}
\alpha_{i}(x,t)= \frac{w_{i}z_{i}(x,t)}{\sum_{i=1}^{n} w_{i}z_{i}(x,t)}.
\end{equation}
where, $z_{i}(x,t) = exp(-v^{*}(x,t)), v^{*}(x,t) = v^{\pi^{*}}(x,t)$, indicates the feasibility of the policy for the current state, and $w_{i}$ is the fixed weights to determine the goal of the controller.

%%%%%%%%%%%%%%%%%%%%%%%
%% explain the dynamics
%%%%%%%%%%%%%%%%%%%%%%%
In order to derive each of the sub-optimal policies $\pi^{*}_{n}$, we used the iterative linear-quadratic-regulator (iLQR) method as the optimal control \cite{iLQG} for non-linear system dynamics described as follows:
\begin{equation}\label{eq:dynamics}
x(k+1) = f(x(k), u(k)).
\end{equation}
where, $x$ indicates the state and $u$ is the input.
%%%%%%%%%%%%%%%%%%%%%%%%%%%%%
%% Explaine the iLQR method
%%%%%%%%%%%%%%%%%%%%%%%%%%%%%
The problem with optimization is to find the optimal control policy $\pi^{*}$ that minimizes the objective function (\ref{eq:cost_func}).
The control policy is calculated as follows: 
\begin{equation}\label{eq:control_low1}
u(k) \leftarrow u(k) + \delta{u(k)}\Delta{t}
\end{equation}
where, $k$ is the sampling number and $\Delta{t}$ is the sampling time.
Each iteration of the iLQR algorithm begins with a nominal control sequence $\overline{u}(t)$ and the corresponding trajectory $\overline{x}(t)$, and is observed by applying $\overline{u}(t)$ to the dynamics equation (\ref{eq:dynamics}) with $\overline{x}(0)$.
The iLQR method solves the optimization problem for non-linear system by linearizing the system dynamics and with second-order approximation of the objective functions around a nominal control sequence $\overline{u}(k)$ and the corresponding nominal trajectory $\overline{x}(k)$. 
Therefore, the iLQR problem is solved using the Riccati-like equations \cite{iLQG} and the control law $\delta u(k)$ is
\begin{equation}\label{eq:control_low2}
\delta u(k) = l(k) + L(k)\delta x(k),
\end{equation}
where, $\delta x(k) \triangleq x(k)-\overline{x}(k)$, and $l(k)=-H^{-1}(k)g(k)$ is an open loop term and $L(k)=-H(k)^{-1}G(k)$ is the feedback term ( see \cite{iLQG}  for further details).
%The process of determining $l$ and $L$ are shown in Appendix.
%Please refer to \cite{iLQG} for the detailed process for determining $l$ and $L$.

%%%%%%%%%%%%%%%%%%%%%%%
%% Experimental setups
%%%%%%%%%%%%%%%%%%%%%%%
\section{Experimental setups}
To validate our approach, we conducted experiments with three healthy right handed subjects, after obtaining informed consent from them.

%%%%%%%%%%%%%%%%%%%%%%%
% Upper limb-exoskeleton 
%%%%%%%%%%%%%%%%%%%%%%%
\subsection{Upper limb exoskeleton robot}
\begin{figure}[t]
	\includegraphics[width=80mm]{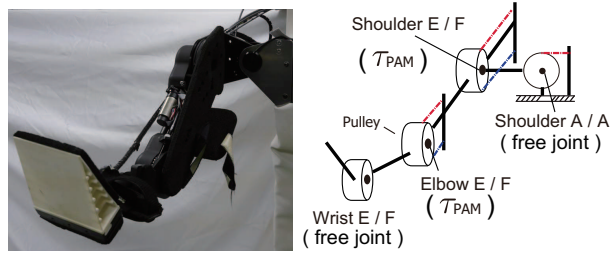}
	\begin{center}
		\caption{Upper-limb exoskeleton robot. Shoulder extension/flexion (SFE) and elbow extension/flexion (EFE) are actuated by pneumatic artificial muscle (PAM) actuators. In this experiment, shoulder abduction/adduction and wrist extension/flexion joints were defined as the free joints.}
		\label{fig:exo}
	\end{center}
\end{figure}
In our robot experiment, the optimal policies were derived to control two joints of the upper limb of the exoskeleton robot (see, Fig.\ref{fig:exo}): shoulder flexion/extension (SFE), elbow flexion/extension(EFE) joints.
Each joint is actuated by pneumatic artificial muscle (PAM).
As depicted in Fig.\ref{fig:exo}, the robot has the remaining two degrees of freedom; shoulder abduction/adduction (SAA) and wrist flexion/adduction (SAA) joints, but in this experiment they were set as free joints.
The joint torques are generated by a PAM as follows:
\begin{equation}
\tau_{pam} = rf_{pam}
\end{equation} 
where, $r$ is the pulley radius and $f_{pam}$ is the PAM force generated by the path contraction of the spiral fibers embedded in a pneumatic bladder. 
The details of the mechanical design and the pressure-force model were introduced in our previous studies \cite{teramaeICRA2014, nodaIROS2014, furukawaTRO}.

In this study, we considered the dynamics of the two-link robot model.
When deriving the control policy to assist the user's motion, we considered the dynamics as a system of human and robot as follows: the user's link weights where the center of the link mass (calculated from the subject's body parameter \cite{biomech}) were added to the robot link weights.
The state variable included the joint angle $\theta$, angular velocity $\dot{\theta}$, and measured inner pressure of PAM $P$ for the dynamics eq.(\ref{eq:dynamics}): 
${\boldsymbol x} = \left[\theta_{1}, \theta_{2}, \dot{\theta_{1}}, \dot{\theta_{2}}, P_{1}, P_{2} \right]^{\top}$.
where, the subscripts $1$ and $2$ at the lower right represent the shoulder and elbow joint, respectively.
The control outputs ${\bf u}$ were the desired pressure inputs for pneumatic actuators $P_{ref}$:
${\boldsymbol u} = \left[P_{ref1}, P_{ref2}\right]^{\top}$.
Although the dynamics between the reference pressure $P_{ref}$ and the measured inner pressure $P$ has non-linearity due to the air dynamics, we can approximate the dynamics by the first order lag system \cite{teramaeICRA2014} as follows:
\begin{equation}
P=\frac{1}{1+t_{c}s}P_{ref}
\end{equation}
where, $t_{c}$ is the time constant and is set as follows:
\[
t_{c} = \begin{cases}
0.08 & (\dot{\tau}_{pam}>0) \\
0.4 & (otherwise)
\end{cases}
\]
In addition, the maximum pressure is set to $0.8$ MPa for safety.

The terminal cost $g(\cdot)$ for minimization was set as follows:
\begin{equation}
\begin{split}
E_{a}(T) &= \left({\boldsymbol \theta}(T)-{\boldsymbol \theta}_{target}(T)\right), \\
E_{v}(T) &= \left(\dot{{\boldsymbol \theta}}(T)-\dot{{\boldsymbol \theta}}_{target}(T)\right), \\
g(x(T)) &= C_{a}E_{a}^{2}(T) + C_{v}E_{v}^{2}(T),
\end{split}
\end{equation}
where, $T$ is the final time, $\theta_{target}$ is the target angles, and $\dot{\theta}_{target}$ is the target angular velocities at the point of release of the ball for the basketball throwing task and was calculated according to goal distance.

The instantaneous cost $l(\cdot)$ was set as
\begin{equation}
l(x,\pi,t) = C_{p}{\boldsymbol P}_{ref}^{2}(t) + C_{pd}\dot{{\boldsymbol P}}_{ref}^{2}(t).
\end{equation}
where, $C_{a}, C_{v}, C_{p}, C_{pd}$ were manually selected as in the most of optimal control studies.
%%%%%%%%%%%%%%%%%%%%%
%% motion task
%%%%%%%%%%%%%%%%%%%%%
\subsection{Motion Task}
\begin{figure}[t]
	\includegraphics[width=80mm]{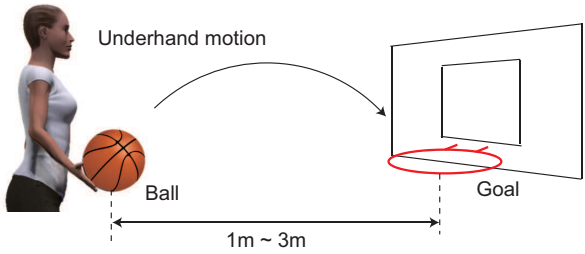}
	\begin{center}
		\caption{Basketball goal shoot task. The subjects throw a basketball with an underhand motion in the sitting condition. During the experiment, the distance of the hoop was changed to 1 m, 2 m, and 3 m, and the subjects are asked to throw the ball under the following two conditions: wearing the exoskeleton robot but No-Assist ({\it NA}), wearing the exoskeleton robot with Assist ({\it A}).}
		\label{fig:task}
	\end{center}
\end{figure}

To evaluate our proposed approach, we set the basketball goal shoot as the concrete motion task as depicted in Fig. \ref{fig:task}.
In this task, the subjects threw a basketball with underhand motion in the sitting condition, and the distance of the hoop was changed to 1 m, 2 m, and 3 m.
The subjects were instructed to throw the ball only with their arms and were asked to throw the ball under two conditions; wearing the exoskeleton robot but No-Assist ({\it NA}), wearing the exoskeleton robot with Assist ({\it A}).
The subjects threw the ball ten times in each condition.
To measure the timing of the release of the ball, one tactile sensor was attached to the tip of the subject's middle finger.

%%%%%%%%%%%%%%%%%%%%%%
% feature selection
%%%%%%%%%%%%%%%%%%%%%%
\subsection{Feature selection and parameter identification based on PLS}
\begin{figure}[t]
	\includegraphics[width=80mm]{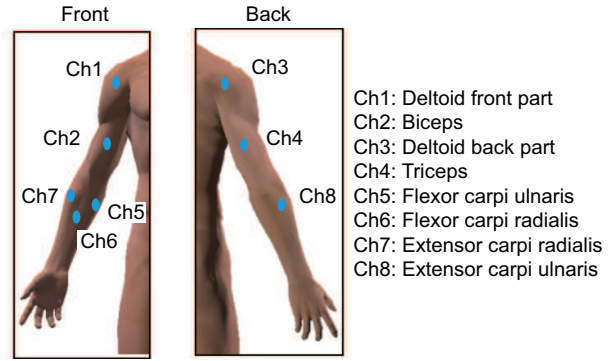}
	\begin{center}
		\caption{EMG channel location. We used 8 channels to estimate the user's motor goal.}
		\label{fig:EMG}
	\end{center}
\end{figure}
In order to estimate the user's motor goal from a slight, initial motion, the rectified and filtered EMG signals: $e$, angles of the shoulder and elbow joint, and angular velocities generated by the shoulder and elbow joint are used as the features: ${\boldsymbol \psi} = \left[e_{1}, e_{2}, e_{3}, e_{4}, e_{5}, e_{6}, e_{7}, e_{8},\theta_{1}, \theta_{2}, \dot{\theta}_{1}, \dot{\theta}_{2}\right]^{\top}$ (totally $12$-dimension).
The EMG signals were measured from the right arms of the subjects.
Figure \ref{fig:EMG} shows the EMG locations probed to measure the muscle activities using $8$ sensor channels (Ch1: $e_{1}$, Ch2: $e_{2}$, Ch3: $e_{3}$, Ch4: $e_{4}$, Ch5: $e_{5}$, Ch6: $e_{6}$, Ch7: $e_{7}$, Ch8: $e_{8}$).
We used Ag/AgCl bipolar surface EMG electrodes.
Using the encoder of the upper-limb exoskeleton robot system, we simultaneously obtained both the angles and angular velocities of the shoulder and elbow joint. 

During the basketball goal shooting motion, the point at which the angular velocity of the shoulder joint exceeds the threshold is defined as the motion start point.
This is because we observed the tendency of the shoulder joint to move earlier than the elbow in this task, from the preliminary experiment.
Moreover, the threshold was set to $0.2$ from the preliminary experiment.
The average value of each data tens of milliseconds before the motion start point is used as the feature value ${\psi}$.
In particular, the EMG signals are activated $60-100$ ms prior to the actual limb movements \cite{kawato1995}, and this latency is set by the cross validation of the training-data.

In this study, the sensor values were acquired as the training data when the exoskeleton robot assisted the motion for shooting the goal at a distance of 1 m and 3 m, assuming that the goal is known.
With the corresponding labes $y = 1$ and $y=3$ ($1$: 1 m throw, $3$: 3 m throw), a projection matrix ${\bf W}$ for calculating the one-dimensional feature $\mu \in \mathcal{R}$ by PLS was derived.
%Furthermore, the goal is estimated from the feature reduced from $12$-dimension to one-dimension.
%%%%%%%%%%%%%%%%%
%% Results
%%%%%%%%%%%%%%%%%
\section{RESULTS}
In this section, we first show the simulation results, and subsequently, the results of the online assist control based on the user's motor goal estimation.

\subsection{Simulation results}
%\begin{figure}[t]
%	\includegraphics[width=85mm]{./figs/simulate_input.eps}
%	\begin{center}
%		\caption{Representative input results derived from the optimal control. (a) Shows the input in the 1 m throw condition, and (b) shows the input in the 3 m throw condition.}
%		\label{fig:simulate_input}
%	\end{center}
%\end{figure}
\begin{figure}[t]
	\includegraphics[width=80mm]{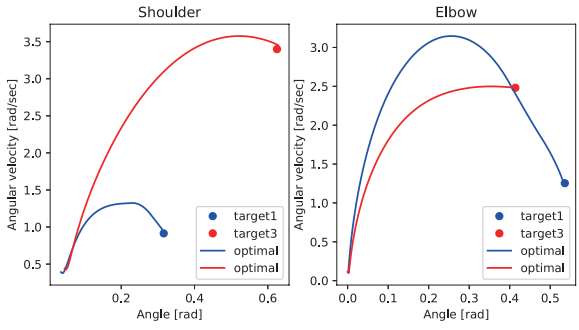}
	\begin{center}
		\caption{Simulation results of the representative state transitions generated by the optimal control policy in 1 m and 3 m throw condition.}
		\label{fig:simulate_state}
	\end{center}
\end{figure}
%Figure \ref{fig:simulate_input} (a) and (b) show the representative input results derived by the optimal control, from the dynamics of one subject wearing a robot for the 1 m throw and 3 m throw.
Figure \ref{fig:simulate_state} shows the simulation results of the state transition generated by the optimal control policies.
Error between the target and generated angles and angular velocities at the release point in final time were as follows:
Shoulder joint angle is $0.003$ rad, angular velocity is $0.034$ rad/s, elbow joint angle is $0.003$ rad, angular velocity is $0.013$ rad/s in the 1 m ball throw condition, whereas, in the 3 m ball throw condition, the shoulder joint angle is $0.0076$ rad, angular velocity is $0.0066$ rad/s, elbow joint angle is $0.0023$ rad, and angular velocity is $0.0036$ rad/s.
Similar results were obtained for dynamics in other users, and moreover, the errors were small.
From these results, we could get the optimal control policy to realize the target state which is necessary for the release point of the basketball.

\begin{figure}[t]
	\includegraphics[width=80mm]{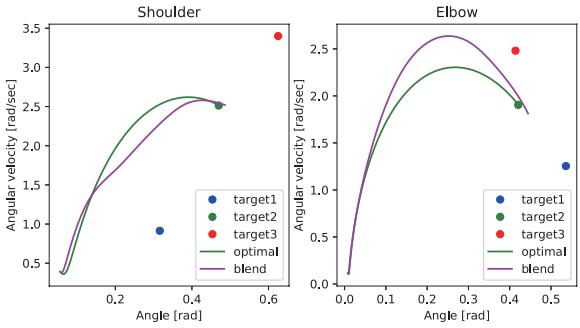}
	\begin{center}
		\caption{State transition results from blended control policy and optimal control policy in 2 m goal shoot condition.}
		\label{fig:blend_state}
	\end{center}
\end{figure}
Figure \ref{fig:blend_state} shows the state transition results from the blending method of the optimal control policy for 2 m goal shot.
To calculate the blended control policy for the 2 m goal shot condition, the weight parameters were set as $w_{1} = 0.5$ and $w_{2} = 0.5$, where, the $w_{1}$ and $w_{2}$ represents the weights for blending the optimal control policy of the 1 m and 3 m goal shot conditions, respectively.
For comparison, the state transition result by optimal control policy for the 2 m goal shot condition is also shown.
Error between the target and generated angles of release point in final time were as follows:
shoulder joint angle, $0.019$ rad; angular velocity, $0.0031$ rad/s; elbow joint angle, $0.0081$ rad and angular velocity, $0.0042$ rad/s, with optimal control policy, and shoulder joint angle, $0.0038$ rad; angular velocity, $0.0067$ rad/s; elbow joint angle, $0.015$ rad, and angular velocity, $0.09$ rad/s, with the blended control policy.
%Although the error of state by the blended control policy was slightly larger than the optimal control policy, 
The final state was close to the target. The result indicates the usefulness of the blended policy for the new goal shot condition.

%\subsection{Online control performance and tendency of PLS result}
\subsection{Feature extraction and weight determination based on PLS}
%\begin{figure}[t]
%	\includegraphics[width=85mm]{./figs/Angle_and_velocity_only_robot.eps}
%	\begin{center}
%		\caption{we measured the state transition of the upper-limb exoskeleton robot controlled by the optimal control policy, derived only with the robot dynamics in the 1 m and 3 m goal shot conditions. In this condition, no user wore the exoskeleton robot.}
%		\label{fig:opt_only_robo}
%	\end{center}
%\end{figure}
%We first evaluated the control performance of the actual upper-limb exoskeleton robot controlled by the optimal control policy, derived only with the robot dynamics.
%In this validation, the robot was not attached to the user.
%Figure \ref{fig:opt_only_robot} shows the measured state transition of the actual upper-limb exoskeleton robot in the 1 m and 3 m goal shot condition.
%Error between the target and generated angles of release point in final time were as follows:
%Shoulder joint angle, $0.0066$ rad; angular velocity, $0.060$ rad/s; elbow joint angle, $0.021$ rad, and angular velocity, $0.0077$ rad/s in the 1 m ball throwing condition, and shoulder joint angle, $0.013$ rad; angular velocity, $0.14$ rad/s; elbow joint angle, $0.019$ rad, and angular velocity, $0.095$ rad/s, in the 3 m ball throwing condition.
%These results show that the actual upper-limb exoskeleton robot is controlled precisely with control policies.

\begin{figure}[t]
	\includegraphics[width=75mm]{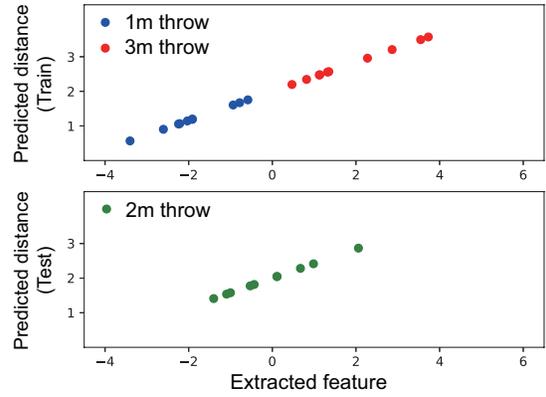}
	\begin{center}
		\caption{Representative tendency of the feature values obtained by PLS and the corresponding user's motor goals.}
		\label{fig:pls}
	\end{center}
\end{figure}
We validated the tendency of the PLS results.
Figure \ref{fig:pls} shows the representative tendency of the feature values of one subject obtained by PLS and the corresponding user's motor goals.
The parameters for PLS were trained by using the 1 m and 3 m goal shot data, and tested with the 2 m goal shot data.
This figure indicates that the feature values and goal labels have order according to the distance condition.
This tendency was seen in other subjects as well.
Therefore, it was shown that the estimated goal label $\hat{y}$ can be used as a user's motor goal calculated by (\ref{eq:weight}).
In this study, based on these results, we set the user's motor goal for the 1 m goal shot as $w_{1} = 1-w_{2}$ and the 3 m goal shot as $w_{2} = \frac{1}{1+exp(-a\hat{y}-b)}$.
If the user intends to throw at a distance of 1 m, the weight $w_{1}$ is high and the $w_{2}$ is small.
Conversely, if the user intends to throw at a distance of 3 m, the weight $w_{1}$ decrease and the $w_{2}$ increases.

\subsection{Online control results of basketball goal shoot}
In this subsection, we describe the experimental results of the basketball goal shooting test with our proposed assist control approach.
%%%%%%%%%%%
% weight ratio
%\begin{figure}[t]
%	\includegraphics[width=80mm]{./figs/weight_ratio.eps}
%	\begin{center}
%		\caption{Average of the user's motor goal estimation results when throwing the ball ten times at each goal distance.}
%		\label{fig:weight_ratio}
%	\end{center}
%\end{figure}
%Figure \ref{fig:weight_ratio} shows the average of the estimated user's motor goals when throwing the ball ten times at each goal distance.
%From this figure, it can be seen that in all the subjects, the ratio of $w_{1}$ is high and $w_{2}$ is small at the 1 m goal shot condition, and the ratio $w_{1}$ is small and $w_{2}$ is high under the 3 m distance goal shot conditions.
%On the contrary, it is shown that the ratio of $w_{1}$ and $w_{2}$ in the 2 m distance goal shot is intermediate compared to that of the 1 m and 3 m goal shots.
%These results show that the user's motor goal is successfully estimated, and the corresponding control policies were derived by blending to control the exoskeleton robot.

%%%%%%%%%%%%%
% EMG
%%%%%%%%%%%%%
\begin{figure}[t]
	\includegraphics[width=85mm]{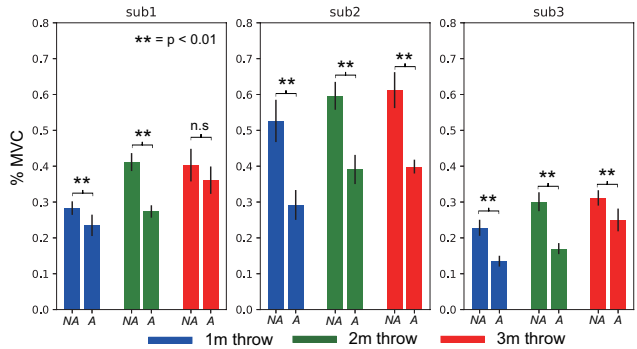}
	\begin{center}
		\caption{Average and standard deviation of MVC for all EMG signals thrown 10 times.}
		\label{fig:emg_average}
	\end{center}
\end{figure}
To verify the effect of the assist, we observed the amplitudes of the rectified and filtered EMG signals.
Figure. \ref{fig:emg_average} shows the average and standard deviation $\%$MVC ($\%$Maximum Voluntary Contraction) of all EMG signals generated during the 10 times the ball was thrown.
The $\%$MVC was the normalized muscle activity for each muscle separately, by using the maximum value of the rectified and low-pass filtered EMG signal: $\%$MVC = $e / e_{max}$.
We applied the Welch's $t$-test between the MVC of {\it NA} and {\it A} conditions.
Among most of the subjects, we found that the average MVCs significantly decreased under condition {\it A} as compared to condition {\it NA}.
Although the 3 m distance goal shot with condition {\it A} in subject 1 did not show a significant difference, it can be confirmed that the average MVCs tended to decrease compared to condition {\it NA}.
These results show that the control policies derived by our proposed approach successfully assist the user's ball throwing motions.

%%%%%%%%%%%%%%%%%%%%%%%%%%%%%%%%%
\begin{table}[t]
\caption{Goal shoot rate of each condition}
\label{table}
\begin{center}
\begin{tabular}{|c|c|c|}
	\hline
	&\multicolumn{2}{|c|}{sub1} \\ 
\hline
& {\it NA} & {\it A}  \\ 
\hline
1m throw & $100$\% & $100$\% \\
\hline
2m throw & $100$\% & $50$\% \\
\hline
3m throw & $0$\% & $50$\% \\
\hline
\hline
& \multicolumn{2}{|c|}{sub2} \\
\hline
& {\it NA} & {\it A} \\
\hline
1m throw & $90$\% & $100$\% \\
\hline
2m throw & $0$\% & $100$\% \\
\hline
3m throw & $0$\% & $0$\% \\
\hline
\hline
 & \multicolumn{2}{|c|}{sub3} \\
 \hline
 & {\it NA} & {\it A} \\
 \hline
1m throw & $100$\% & $80$\% \\ 
\hline
2m throw & $10$\% & $60$\% \\
\hline
3m throw & $0$\% & $70$\% \\
\hline
\end{tabular}
\end{center}
\end{table}
%%%%%%%%%%%%%%%%%%%%%%%%%%%%%%%%%%
%% hoop hit rate
%%%%%%%%%%%%%%%%%%%%%%%%%%%%%%%%%%
%\begin{table}[t]
%	\caption{Number of goal hoop hits in failed shoots (number of hits / number of failures)}
%	\label{table_foop}
%	\begin{center}
%		\begin{tabular}{|c|c|c|c|}
%			\hline
%			&\multicolumn{3}{|c|}{sub1} \\ 
%			\hline
%			& NR & NA & A  \\ 
%			\hline
%			1m throw & No-failure & No-failure & No-failure \\
%			\hline
%			2m throw & $2/2$ & No-failure & $5/5$ \\
%			\hline
%			3m throw & $6/7$ & $0/10$ & $5/5$ \\
%			\hline
%			\hline
%			& \multicolumn{3}{|c|}{sub2} \\
%			\hline
%			& NR & NA & A \\
%			\hline
%			1m throw & No-failure & $1/1$ & No-failure \\
%			\hline
%			2m throw & $1/1$ & $0/10$ & No-failure \\
%			\hline
%			3m throw & $0/10$ & $0/10$ & $0/10$ \\
%			\hline
%			\hline
%			& \multicolumn{3}{|c|}{sub3} \\
%			\hline
%			1m throw & $1/1$ & No-failure & $2/2$ \\ 
%			\hline
%			2m throw & $3/3$ & $6/9$ & $4/4$ \\
%			\hline
%			3m throw & $5/7$ & $0/10$ & $2/3$ \\
%			\hline
%		\end{tabular}
%	\end{center}
%\end{table}
%%%%%%%%%%%%%%%%%%%%%%%%%%%%%%%%%%%%%%%%
Table \ref{table} shows the goal shoot rate when throwing the ball ten times in each of the conditions, by each subject.
When throwing the ball at a distance of 1 m, the shoot rate under the {\it NA} and {\it A} conditions did not change much for all subjects, And therefore, It was considered that the goal distance was close and easy task.
However, it was observed that the assist control did not disturb the ball throwing motion. 
This was the result of correct estimation of the user's motor goal.
At a goal distance of 2 m, the subjects 2 and 3 had higher shoot rate as compared to the {\it NA} condition.
This indicates that the goal shoot can be made with high accuracy and less effort in condition {\it A} as compared to the {\it NA} condition.
On the contrary, in the subject 1, although the user's motor goal was accurately estimated, the shoot accuracy was lower in condition {\it A} than in {\it NA} condition.
This might be because the 2 m goal shoot distance would have been an easy task for this user.
Therefore, although the exoskeleton robotic assistance reduced the effort of the motion, it may not have affected the goal shoot accuracy.
At a goal distance of 3 m, the subjects 1 and 3 demonstrated that the shoot rate under condition {\it A} is higher compared to that in condition {\it NA} with lower effort.
With respect to subject 2, the shoot rate remained at $0\%$ even under condition {\it A}, and this goal distance task seems to be quite difficult even though the robot assisted the motion.

From these results, it was shown that our proposed approach is effective in assisting the control for motion that requires speed and accuracy.

To further evaluate the shoot rate with our proposed approach, it can be beneficial to explicitly take the control of other joints, e.g., shoulder abduction/adduction, wrist joint, into account. In addition, in this research, the relationship between the feature values and user’s motor goals is ordered, extending our method to cope with the tasks where the relationship is complicated would be one of the interesting directions as a part of future study.

%%%%%%%%%%%%%%%%%%%
%%
%% Conclusions
%%
%%%%%%%%%%%%%%%%%%%
\section{CONCLUSIONS}
Our findings indicated that the proposed approach for controlling the assistive exoskeleton robot was successful in the simulated and actual basketball shooting experiments. The effort required to throw the ball was lower under the assisted condition than under the other conditions, with accuracy increasing as task difficulty increased in most participants. 
These results demonstrate that our proposed approach effectively allows for robot-assisted control of motions requiring speed and accuracy. 

%%%%%%%%%%%%%%%%%%%%%%%%%%%%%%%%%
%% Mention of the Limitation2
%%%%%%%%%%%%%%%%%%%%%%%%%%%%%%%%%
In this study, control policies were blended using a linear Bellman combination approach. That is, control policies for the 2-m condition were derived based on the optimal control policies for the 1-m and 3-m conditions. Our findings suggest that this approach can be used for the coordination of new tasks \cite{LBC_silva}, although further studies are required to examine this hypothesis. Future studies should also consider extending this approach to handle a variety of motions that require greater degrees of freedom.

\section*{ACKNOWLEDGMENT}
This work was supported by JST, ACT-i, Grant Number JPMJPR18UQ, Japan.
The results presented have been achieved by "Research and development of technology for enhancing functional recovery of elderly and disabled people based on non-invasive brain imaging and robotic devices," the commissioned research of the National Institute of Information and Communications Technology (NICT), JAPAN.
Part of this research was supported by JSPS KAKENHI Grant Number JP18K18135.
\bibliographystyle{IEEEtran}
\bibliography{opt_assist_ref_v4}
%\bibliography{test}
\end{document}